\documentclass[dvipsnames,format=sigconf]{acmart}

\copyrightyear{2024}
\acmYear{2024}
\setcopyright{rightsretained}
\acmConference[GECCO '24 Companion]{Genetic and Evolutionary Computation Conference}{July 14--18, 2024}{Melbourne, VIC, Australia}
\acmBooktitle{Genetic and Evolutionary Computation Conference (GECCO '24 Companion), July 14--18, 2024, Melbourne, VIC, Australia}
\acmDOI{10.1145/3638530.3664090}
\acmISBN{979-8-4007-0495-6/24/07}

\usepackage{adjustbox}
\usepackage{algorithmic}
\usepackage{amsmath,amsfonts}
\usepackage{array}
\usepackage{bibunits}
\usepackage{booktabs}
\usepackage{caption}
\usepackage{cite}
\usepackage{colortbl}
\usepackage{csvsimple}
\usepackage{csquotes}
\usepackage{etoolbox}
\usepackage{graphicx}
\usepackage{hyperref,cleveref}
\usepackage{import}
\usepackage{rotating}
\usepackage[subtle]{savetrees}
\usepackage{siunitx}
\usepackage{subcaption}
\usepackage{tabularx}
\usepackage{textcomp}
\usepackage{xcolor}
\usepackage{orcidlink}
\usepackage{pdflscape,longtable}
\usepackage{listings}
\usepackage{placeins}
\usepackage{pythonhighlight}

\makeatletter
\let\pragma@iinput=\@iinput
\def\@iinput#1{\xdef\@pragmafile{#1}\pragma@iinput{#1} }
\def\@pragmafile{default}
\def\pragmaonce{%
   \csname pragma@\@pragmafile\endcsname
   \global\expandafter\let \csname pragma@\@pragmafile\endcsname =  
}
\makeatother

\ifdefined\mydraft
\mydraft
\fi

\begin{document}

\title{Trackable Island-model Genetic Algorithms at Wafer Scale}

\author{Matthew Andres Moreno}
\orcid{0000-0003-4726-4479}
\author{Connor Yang}
\orcid{0009-0004-1240-2362}
\affiliation{%
  \institution{University of Michigan}
  \city{Ann Arbor}
  \state{Michigan}
  \country{USA}
}

\author{Emily Dolson}
\orcid{0000-0001-8616-4898}
\affiliation{%
  \institution{Michigan State University}
  \city{East Lansing}
  \state{Michigan}
  \country{USA}
}

\author{Luis Zaman}
\orcid{0000-0001-6838-7385}
\affiliation{%
  \institution{University of Michigan}
  \city{Ann Arbor}
  \state{Michigan}
  \country{USA}
}

\renewcommand{\shortauthors}{Moreno et al.}


\begin{abstract}
Emerging ML/AI hardware accelerators, like the 850,000 processor Cerebras Wafer-Scale Engine (WSE), hold great promise to scale up the capabilities of evolutionary computation.
However, challenges remain in maintaining visibility into underlying evolutionary processes while efficiently utilizing these platforms' large processor counts.
Here, we focus on the problem of extracting phylogenetic information from digital evolution on the WSE platform.
We present a tracking-enabled asynchronous island-based genetic algorithm (GA) framework for WSE hardware.
Emulated and on-hardware GA benchmarks with a simple tracking-enabled agent model clock upwards of 1 million generations a minute for population sizes reaching 16 million.
This pace enables quadrillions of evaluations a day.
We validate phylogenetic reconstructions from these trials and demonstrate their suitability for inference of underlying evolutionary conditions.
In particular, we demonstrate extraction of clear phylometric signals that differentiate wafer-scale runs with adaptive dynamics enabled versus disabled.
Together, these benchmark and validation trials reflect strong potential for highly scalable evolutionary computation that is both efficient and observable.
Kernel code implementing the island-model GA supports drop-in customization to support any fixed-length genome content and fitness criteria, allowing it to be leveraged to advance research interests across the community.

\end{abstract}



\maketitle

\section{Introduction}

\begin{figure}
  \vspace{2ex}
    \centering
  \includegraphics[width=0.8\linewidth]{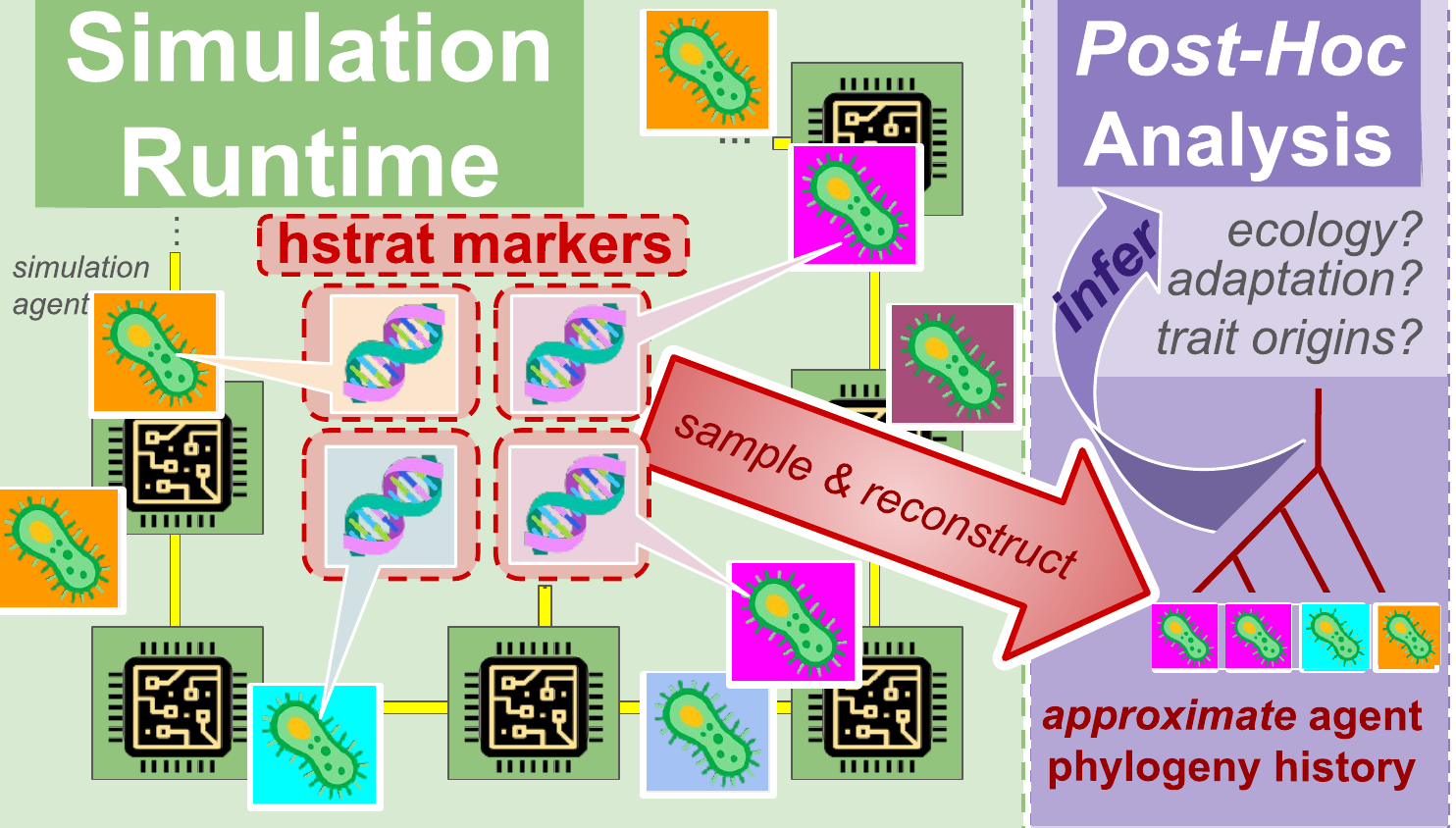}
  \vspace{-1.5ex}
    \caption{\textbf{Strategy for trackable distributed evolution simulation.}
    \footnotesize %
    Hstrat markers attached to agents support post hoc estimation of relatedness between lineages, enabling approximate phylogenetic reconstruction.
    }
    \label{fig:runtime-posthoc-schematic}
\vspace{-0.2in}
\end{figure}

A major upshot of the deep learning race is the emergence of spectacularly capable next-generation compute accelerators \citep{zhang2016cambricon,emani2021accelerating,jia2019dissecting,medina2020habana}.
Although tailored expressly to deep learning workloads, these hardware platforms represent an exceptional opportunity to springboard the capabilities of evolutionary computation.
The emerging class of fabric-based accelerators, led by the 850,000 core Cerebras CS-2 Wafer Scale Engine (WSE) \citep{lauterbach2021path}, holds particular promise in this regard.
This architecture interfaces multitudinous processor elements (PEs) in a physical lattice, with PEs executing independently and interacting locally through a network-like interface.

Our work here explores how such hardware might be recruited for large-scale, telemetry-enabled digital evolution, demonstrating a GA implementation tailored to the dataflow-oriented computing model of the CS-2 platform.
We apply hereditary stratigraphy algorithms to enable \textit{post hoc} reconstruction of phylogenetic history from special fitness-neutral annotations on genomes (Figure \ref{fig:runtime-posthoc-schematic}) \citep{moreno2022hereditary} .

\section{Methods} \label{sec:methods}

We apply an island-model genetic algorithm, common in applications of parallel and distributed computing to evolutionary computation, to instantiate a wafer-scale evolutionary process spanning the Wafer-Scale Engine's PEs.
Under this model, PEs each host independent populations and exchange genomes with neighbors.

Our implementation unfolds according to a generational update cycle, with migration handled first.
Each PE maintains independent immigration buffers and emigration buffers dedicated to each cardinal neighbor.
Asynchronous operations are registered to on-completion callbacks that set a per-buffer ``send-'' or ``receive-complete'' flag variable.
The main population buffer held 32 genomes.

Subsequently, the main update loop tests all completion flags.
For each immigration flag that is set, buffered genomes are copied into the main population buffer, replacing randomly chosen population members.
Likewise, for each emigration flag set, corresponding send buffers are re-populated with randomly sampled genomes from the main population buffer.
Corresponding flags are then reset and new async requests are initiated.

The remainder of the main update loop handles evolutionary operations within the scope of the executing PE.
Each genome within the population is evaluated to produce a scalar fitness value.
After evaluation, tournament selection is applied.
Each slot in the next generation is populated with the highest-fitness genome among $n=5$ randomly sampled individuals, with ties broken randomly.

Finally, a mutational operator is applied across all genomes in the next population.
At this point, hereditary stratigraphy annotations are updated to reflect an elapsed generation.
The next generation cycle then launches, unless a generation count halting condition is met.




Kernel code for the island-model GA is open source at\\ \href{https://github.com/mmore500/wse-sketches/tree/v0.2.2}{\texttt{github.com/mmore500/wse-sketches/tree/v0.2.2}}, and can be harnessed for any fixed-length genome model and evaluation criteria by drop-in replacement of one source file containing three functions (initialize, mutate, and evaluate).
We hope to see this work contribute to scale-up in use cases across the research community.
We used Cerebras SDK v1.0.0 \citep{selig2022cerebras}, which allows code to be developed and tested regardless of access to Cerebras hardware.
Hereditary stratigraphy utilities were used from the \textit{hstrat} Python package \citep{moreno2022hstrat}.
For more details, see our upcoming publication on this work \citep{moreno2024trackable}.

\section{Results and Discussion} \label{sec:results}

\begin{figure}

\begin{subfigure}[c]{\linewidth}
  \centering
    \includegraphics[width=0.9\linewidth]{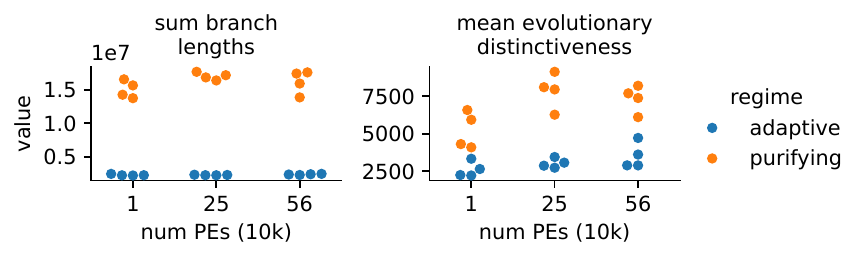}
 \caption{\footnotesize adaptive vs. purifying phylometric structure}
  \label{fig:on-device-phylometrics}
\end{subfigure}

\vspace{0.2ex}

\begin{subfigure}[c]{0.5\linewidth}
  \centering
  \includegraphics[width=0.7\linewidth,trim={0 11.05in 7.2in 0in},clip]{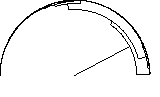}

  \vspace{-2ex}
  \includegraphics[width=0.7\linewidth,trim={0 11.05in 7.2in 0in},clip]{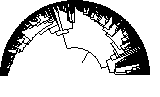}

  \vspace{-2ex}
  \footnotesize
 \caption{\footnotesize adaptive regime}
  \label{fig:on-device-adaptive}
\end{subfigure}%
\begin{subfigure}[c]{0.5\linewidth}
  \centering
  \includegraphics[width=0.7\linewidth,trim={0 11.05in 7.2in 0in},clip]{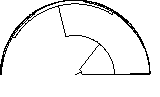}

  \vspace{-2ex}
  \includegraphics[width=0.7\linewidth,trim={0 11.05in 7.2in 0in},clip]{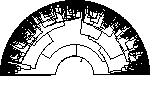}

  \vspace{-2ex}
  \footnotesize
 \caption{\footnotesize purifying regime}
  \label{fig:on-device-purifying}
\end{subfigure}

\vspace{-1.5ex}

\caption{%
\textbf{On-hardware Trial.}
\footnotesize
Results from phylogenetic reconstruction of 1 million generation on-hardware simulations.
Panel \ref{fig:on-device-phylometrics} compares phylometric readings from purifying-only and adaptation-enabled configurations, 4 replicates each.
Panels \labelcref{fig:on-device-adaptive,fig:on-device-purifying} juxtapose example $750\times750$ PE simulation phylogenies under each configuration regime.
Phylometrics were calculated from reconstructions with 10k sampled end-state agents.
For legibility, phylogeny visualizations were further subsampled to 1k end-state agents.
Top phylogenies are linear-scaled.
Bottom phylogenies are log-scaled with ultrametric correction to better show topology.
}
\label{fig:on-device}
\vspace{-0.2in}
\end{figure}

For on-hardware proof-of-concept experiments, we used a 128-bit, tracking-enabled genome layout, with the full first 32 bits containing a floating point fitness value.
We defined two treatments: \textit{purifying-only}, where 33\% of agent replication events decreased fitness by a normally-distributed amount,  and \textit{adaption-enabled}, which added beneficial mutations that increased fitness by a normally-distributed amount, occurring with 0.3\% probability.
These beneficial mutations introduced the possibility for selective sweeps.
As before, we used tournament size 5 for both treatments.
We performed four on-hardware replicates of each treatment instantiated on 10k ($100\times100$), 250k ($500\times500$) and 562.5k ($750\times750$) PE arrays.
We halted each PE after it elapsed 1 million generations.

Across eight on-device, tracking-enabled trials of 1 million generations, we measured a mean simulation rate of 17,688 generations per second for 562,500 PEs ($750\times750$ rectangle) with run times slightly below one minute.
Trials with 1,600 PEs ($40\times40$) performed similarly, completing 17,734 generations per second.

Multiplied out to a full day, 17 generations per second turnover would elapse around 1.5 billion generations.
With 32 individuals hosted per each of 850,000 PEs, the net population size would sit around 27 million at full CS-2 wafer scale.
(Note, though, that available on-chip memory could support thousands of our very simple agents per PE, raising the potential for a net population size on the order of a billion agents.)
A naive extrapolation estimates on the order of a quadrillion agent replications could be achieved hourly at full wafer scale for such a very simple model.
We look forward to more benchmarking and scaling experiments in future work.

Upon completion, we sampled one genome from each PE.
Then, we performed an agglomerative trie-based reconstruction from subsamples of 10k end-state genomes \citep{moreno2024analysis}.
Figure \ref{fig:on-device} compares phylogenies generated under the purifying-only and adaption-enabled treatments.
As expected \citep{moreno2023toward}, purifying-only treatment phylogenies consistently exhibited greater sum branch length and mean evolutionary distinctiveness, with the effect on branch length particularly strong.
These structural effects are apparent in example phylogeny trees from 562.5k PE trials (Figures \labelcref{fig:on-device-adaptive,fig:on-device-purifying}).
Successful differentiation between treatments is a highly promising outcome.
This result not only supports the correctness of our methods and implementation, but also confirms the capability of reconstruction-based analyses to meaningfully describe very large-scale evolving populations.

\begin{acks}
Supported in part by the Schmidt AI in Sci Fellowship.
Thank you also to M. Jacquelin, U. Mody, and L. Wilson at Cerebras Systems.
\end{acks}

\bibliographystyle{ACM-Reference-Format}
\bibliography{bibl}







\end{document}